\documentclass[10pt,twocolumn,letterpaper]{article}

\renewcommand\footnotemark{}

\usepackage{iccv}
\usepackage{times}
\usepackage{epsfig}
\usepackage{graphicx}
\usepackage{amsmath}
\usepackage{amssymb}
\usepackage{color}
\usepackage{subfigure}
\usepackage{tabularx}
\usepackage{multirow}
\usepackage{pifont}

\newcommand{\cmark}{\ding{51}}%
\newcommand{\xmark}{\ding{55}}%

\newcommand{\figref}[1]{Fig. \ref{#1}}
\newcommand{\tabref}[1]{Table \ref{#1}}

\makeatletter
\def\hlinewd#1{%
	\noalign{\ifnum0=`}\fi\hrule \@height #1 \futurelet
	\reserved@a\@xhline}

\usepackage[breaklinks=true,bookmarks=false]{hyperref}

\iccvfinalcopy 

\def\iccvPaperID{5700} 

\ificcvfinal\fi

\begin{document}

\title{Context-Aware Emotion Recognition Networks\thanks{This research was supported by Next-Generation Information Computing Development Program through the National Research Foundation of Korea(NRF) funded by the Ministry of Science and ICT (NRF-2017M3C4A7069370).}}
\author{
	Jiyoung Lee$^{1}$, Seungryong Kim$^{2}$, Sunok Kim$^{1}$, Jungin Park$^{1}$, Kwanghoon Sohn$^{1,*}$\thanks{$^{*}$Corresponding author}\\
	$^1$Yonsei University, $^2$\'Ecole Polytechnique F\'ed\'erale de Lausanne (EPFL)\\
	{\tt\small \{easy00,kso428,newrun,khsohn\}@yonsei.ac.kr, seungryong.kim@epfl.ch}\\
}

\maketitle
\ificcvfinal\thispagestyle{empty}\fi

\begin{abstract}
Traditional techniques for emotion recognition have focused on the facial expression analysis only, thus providing limited ability to encode context that comprehensively represents the emotional responses.
We present deep networks for context-aware emotion recognition, called CAER-Net, that exploit not only human facial expression but also context information in a joint and boosting manner.
The key idea is to hide human faces in a visual scene and seek other contexts based on an attention mechanism. Our networks consist of two sub-networks, including two-stream encoding networks to separately extract the features of face and context regions, and adaptive fusion networks to fuse such features in an adaptive fashion. We also introduce a novel benchmark for context-aware emotion recognition, called CAER, that is more appropriate than existing benchmarks both qualitatively and quantitatively.
On several benchmarks, CAER-Net proves the effect of context for emotion recognition.
Our dataset is available at \small{\url{http://caer-dataset.github.io}}.
\end{abstract}

\section{Introduction}\label{sec:1}
Recognizing human emotions from visual contents has attracted significant attention in numerous computer vision applications such as health care and human-computer interaction systems~\cite{d2007toward,lisetti2003developing,yannakakis2011experience}.
\begin{figure}
	\centering
		\renewcommand{\thesubfigure}{}
		\subfigure{\includegraphics[width=0.33\linewidth]{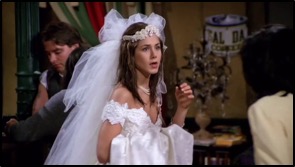}}\hfill
		\subfigure{\includegraphics[width=0.33\linewidth]{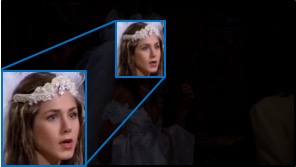}}\hfill
		\subfigure{\includegraphics[width=0.33\linewidth]{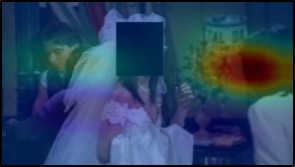}}\hfill \\ \vspace{-10pt}
		\subfigure{\includegraphics[width=0.33\linewidth]{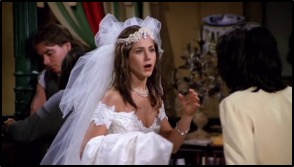}}\hfill
		\subfigure{\includegraphics[width=0.33\linewidth]{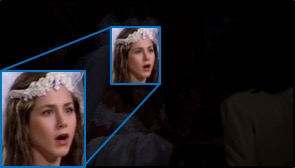}}\hfill
		\subfigure{\includegraphics[width=0.33\linewidth]{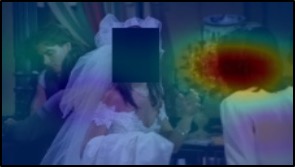}}\hfill \\ \vspace{-10pt}
		\subfigure[(a)]{\includegraphics[width=0.33\linewidth]{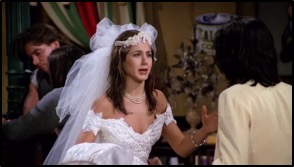}}\hfill
		\subfigure[(b)]{\includegraphics[width=0.33\linewidth]{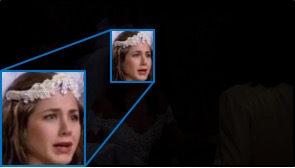}}\hfill
		\subfigure[(c)]{\includegraphics[width=0.33\linewidth]{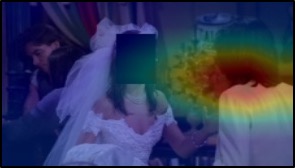}}\hfill \\ \vspace{-3pt}
		\caption{Intuition of CAER-Net: for untrimmed videos as in (a), conventional methods that leverage the facial regions only as in (b) often fail to recognize emotion. Unlike these methods, CAER-Net focuses on both face and attentive context regions as in (c).}
			\label{fig:1}\vspace{-10pt}
\end{figure}

Previous researches for emotion recognition based on handcrafted features~\cite{shan2009facial,zhong2012learning} or deep networks~\cite{fabian2016emotionet,li2018occlusion, li2017reliable} have mainly focused on the perception of the facial expression, based on the assumption that facial images are one of the most discriminative features of emotional responses. In this regard, the most widely used datasets, such as the AFEW~\cite{dhall2011acted} and FER2013~\cite{goodfellow2013challenges}, only provide the cropped and aligned facial images.
However, those conventional methods with the facial image dataset frequently fail to provide satisfactory performance when the emotional signals in the faces are indistinguishable and ambiguous. Meanwhile, people recognize the emotion of others from not only their faces but also surrounding contexts, such as action, interaction with others, and place~\cite{barrett2011context,aminoff2013role}.
Given untrimmed videos as in~\figref{fig:1}(a), could we catch how a woman feels solely from her facial expression as in~\figref{fig:1}(b)?
It is ambiguous to estimate the emotion only with cropped facial videos. However, we could easily guess the emotion as ``surprise" with her facial expression and contexts that an another woman comes close to her as shown in \figref{fig:1}(c). Nevertheless, such contexts have been rarely considered in most existing emotion recognition methods and benchmarks.

Some methods~\cite{chen2016emotion, kosti2017emotion} have shown that emotion recognition performance can be significantly boosted by considering context information such as gesture and place~\cite{chen2016emotion, kosti2017emotion}.
In addition, in visual sentimental analysis~\cite{li2012context,yang2018weakly} that recognizes the sentiment of an image, similar to emotion recognition but not tailored to humans, the holistic visual appearance was used to encode such contexts.
However, these approaches are not practical for extracting the salient context information from visual contents. Moreover, large-scale emotion recognition datasets, including various context information close in real environments, are absence.

To overcome these limitations, we present a novel framework, called Context-Aware Emotion Recogntion Networks (CAER-Net), to recognize human emotion from images and videos by exploiting not only human facial expression but also scene contexts in a joint and boosting manner, instead of only focusing on the facial regions as in most existing methods~\cite{shan2009facial,zhong2012learning,fabian2016emotionet,li2018occlusion,li2017reliable}. The networks are designed in a two-stream architecture, including two feature encoding stream; face encoding and context encoding streams.
Our key ingradient is to seek other relevant contexts by hiding human faces based on an attention mechanism, which enables the networks to reduce an ambiguity and improve an accuracy in emotion recognition.
The face and context features are then fused to predict the emotion class in an adaptive fusion network by inferring an optimal fusion weight among the two-stream features.

In addition, we build a novel database, called Context-Aware Emotion Recognition (CAER), by collecting a large amount of video clips from TV shows and annotating the ground-truth emotion category.
Experimental results show that CAER-Net outperforms baseline networks for context-aware emotion recognition on several benchmarks, including AFEW~\cite{dhall2011acted} and our CAER dataset.

\section{Related Work}\label{sec:2}
\paragraph{Emotion recognition approaches.}\label{sec:21}
Most approaches to recognize human emotion have focused on facial expression analysis~\cite{shan2009facial,zhong2012learning,fabian2016emotionet,li2018occlusion,li2017reliable}. Some methods are based on the facial action coding system ~\cite{friesen1978facial,eleftheriadis2015discriminative}, where a set of localized movements of the face is used to encode facial expression.
Compared to conventional methods that have relied on handcrafted features and shallow classifiers~\cite{shan2009facial,zhong2012learning}, recent deep convolutional neural networks (CNNs) based approaches have made significant progress~\cite{fabian2016emotionet}.
Various techniques to capture temporal dynamics in videos have also been proposed making connections across the time using recurrent neural networks (RNNs) or deep 3D-CNNs~\cite{fan2016video,lee2018spatiotemporal}.
However, most works have been relied on human face analysis, and thus they have limited ability to exploit context information for emotion recognition in the wild.

To solve these limitations, some approaches using other visual clues have been proposed~\cite{nicolaou2011continuous, schindler2008recognizing, chen2016emotion, kosti2017emotion}.
Nicolaou~\etal~\cite{nicolaou2011continuous} used the location of shoulders and Schindler \etal~\cite{schindler2008recognizing} used the body pose to recognize six emotion categories under controlled conditions.
Chen \etal~\cite{chen2016emotion} detected events, objects, and scenes using pre-learned CNNs and fused each score with context fusion.
In~\cite{kosti2017emotion}, manually annotated body bounding boxes and holistic images were leveraged.
However, \cite{kosti2017emotion} have a limited ability to encode dynamic signals (\ie, video) to estimate the emotion.
Moreover, the aforementioned methods are a lack of practical solutions to extract the sailent context information and exploit it to context-aware emotion recognition.

\begin{figure*}
	\centering
	\renewcommand{\thesubfigure}{}
	\subfigure[]{	\includegraphics[width=1\textwidth]{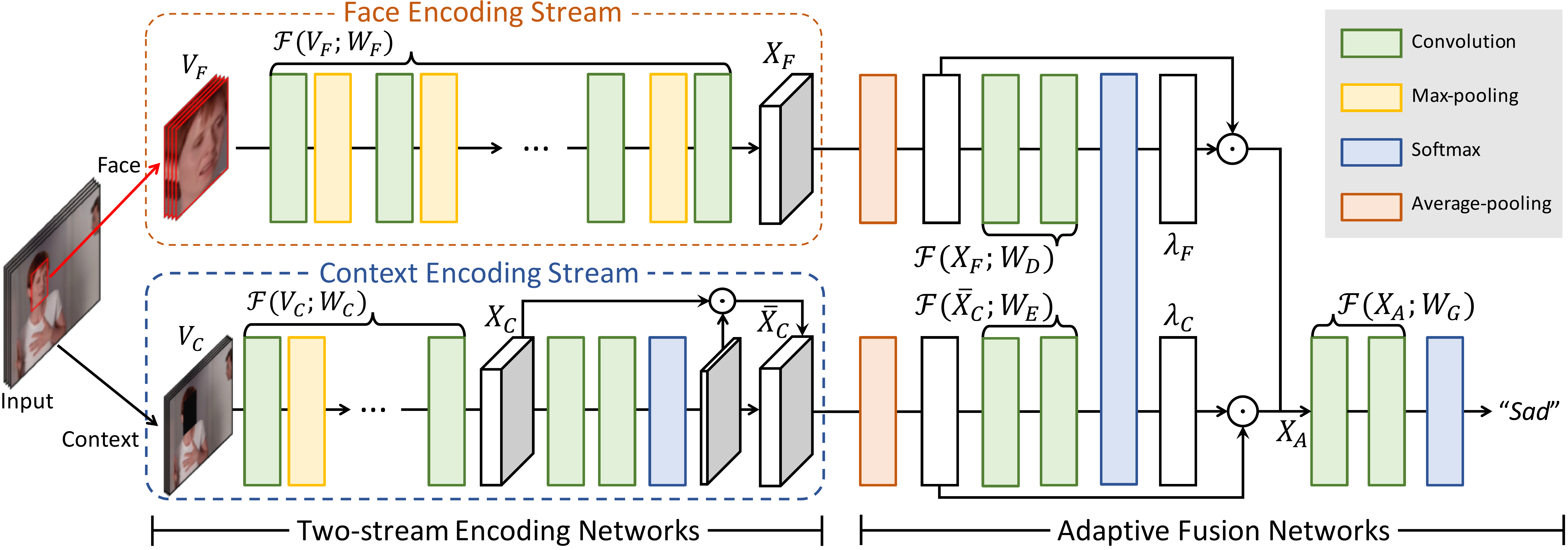}}\\
	\vspace{-10pt}
	\caption{Network configuration of CAER-Net, consisting of two-stream encoding networks and adaptive fusion networks.}
	\label{fig:2}\vspace{-10pt}
\end{figure*}

\vspace{-10pt}
\paragraph{Emotion recognition datasets.}\label{sec:22}
Most of the datasets that focus on detecting occurrence of expressions, such as CK+~\cite{lucey2010extended} and MMI~\cite{pantic2005web}, have been taken in lab-controlled environments. Recently, datasets recorded in the wild condition for including naturalistic emotion states~\cite{dhall2011acted,dhall2011static,mollahosseini2016facial} have attracted much attention. AFEW benchmark~\cite{dhall2011acted} of the EMOTIW challenge~\cite{dhall2016emotiw} provides video frames extracted from movies and TV shows, while SFEW database~\cite{dhall2011static} has been built as a static subset of the AFEW.
FER-Wild~\cite{mollahosseini2016facial} database contains 24,000 images that are obtained by querying emotion-related terms from search engines.
MS-COCO database~\cite{patterson2016coco} has been recently annotated with object attributes, including some emotion categories for human, but the attributes are not intended to be exhaustive for emotion recognition, and not all people are annotated with emotion attributes.
Some studies~\cite{kleinsmith2007recognizing, kleinsmith2011automatic} built the database consisting of a spontaneous subset acquired under a restrictive setting to establish the relationship between emotion and body posture.
EMOTIC database~\cite{kosti2017emotion} has been introduced providing the manually annotated body regions which contains emotional state.
Although these datasets investigate a different aspect of emotion recognition with contexts, a large-scale dataset for context-aware emotion recognition is absence that contains various context information.

\vspace{-10pt}
\paragraph{Attention inference.}\label{sec:23}
Since deep CNNs have achieved a great success in many computer vision areas~\cite{krizhevsky2012imagenet,simonyan2014very,he2016deep}, numerous attention inference models~\cite{zhou2016learning,selvaraju2017grad} have been investigated to identify discriminative regions where the networks attend, by mining discriminative regions~\cite{kumar2016track},
implicitly analyzing the higher-layer activation maps~\cite{zhou2016learning,selvaraju2017grad}, and designing different architecture of attention modules~\cite{woo2018cbam,hu2018squeeze}.
Although the attention produced by these conventional methods could be used as a prior for various tasks, it only covers most discriminative regions of the object, and thus frequently fails to capture other discriminative parts that can help performance improvement.

Most related methods to our work discover attentive areas for visual sentiment recognition~\cite{yang2018weakly,you2017visual}.
Although those produce the emotion sentiment map using deep CNNs, it only focuses on image-level sentiment analysis, not human-centric emotion like us.

\section{Proposed Method}\label{sec:3}
\subsection{Motivation and Overview}\label{sec:31}
In this section, we describe a simple yet effective framework for context-aware emotion recognition in images and videos that exploits the facial expression and context information in a boosting and synergistic manner.
A simple solution is to use the holistic visual appearance similar to~\cite{kosti2017emotion,chen2016emotion}, but such a model cannot encode salient contextual regions well. Based on the intuition that emotions can be recognized by understanding the context components of scene, as well as facial expression together, we present an attention inference module that estimates the context information in images and videos. By hiding the facial regions in inputs and seeking the attention regions, our networks localize more discriminative context regions that are used to improve emotion recognition accuracy in a context-aware manner.

Concretely, let us denote an image and a video that consists of a sequence of $T$ images as $I$ and $V=\{I_1, \dots, I_T\}$, respectively.
Our objective is to infer the discrete emotion label $y$ among $K$ emotion labels $\{y_1,\dots, y_K\}$ of the image $I$ or video clip $V$ with deep CNNs. To solve this problem, we present a network architecture consisting of two sub-networks, including a \textit{two-stream encoding network} and an \textit{adaptive fusion network}, as illustrated in \figref{fig:2}. The two-stream encoding networks consist of \textit{face stream} and \textit{context stream} in which facial expression and context information are encoded in the separate networks. By combining two features in the adaptive fusion network, our method attains an optimal performance for context-aware emotion recognition.

\subsection{Network Architectures}\label{sec:32}
\subsubsection{Two-stream Encoding Networks}\vspace{-5pt}
In this section, we first present a dynamic model of our networks for analyzing videos, and then present a static model for analyzing images.\vspace{-10pt}

\paragraph{Face encoding stream.}
As in existing facial expression analysis approaches~\cite{fabian2016emotionet,lee2018spatiotemporal, fan2018video}, our networks also have the facial expression encoding module. We first detect and crop the facial regions using the off-the-shelf face detectors~\cite{king2009dlib} to build input of face stream $V_F$. The facial expression encoding module is designed to extract the facial expression features denoted as $X_F$ from temporally stacked face-cropped inputs $V_F$ by feed-foward process such that
\begin{equation}
X_F = \mathcal{F}(V_F; W_F),
\end{equation}
with face stream parameters $W_F$.
The facial expression encoding module is designed based on the basic operations of 3D-CNNs which are well-suited for spatiotemporal feature representation.
Compared to 2D-CNNs, 3D-CNNs have the better ability to model temporal information for videos using 3D convolution and 3D pooling operations.

Specifically, the face encoding module consist of 5 convolutional layers with $3 \times 3 \times 3$ kernels followed by batch normalization (BN), rectified linear unit (ReLU) layers and 4 max-pooling layers with stride $2 \times 2 \times 2$ except for the first layer.
The first pooling layer has a kernel size $1 \times 2 \times 2$ with the intention of not to merge the temporal signal too early.
The number of kernels for five convolution layers are 32, 64, 128, 256 and 256, respectively.
The final feature ${X}_F$ is spatially averaged in the average-pooling layer.\vspace{-10pt}

\paragraph{Context encoding stream.}
In comparison to the face encoding stream, the context encoding stream includes a context encoding module and an attention inference module.
To extract the context information except the facial expression, we present a novel strategy that hides the faces and seeks contexts based on the attention mechanisms.
Specifically, the context encoding module is designed to extract the context features denoted as $X_C$ from temporally stacked face-hidden inputs $V_C$ by feed-foward process:
\begin{equation}
X_C = \mathcal{F}(V_C; W_C),
\end{equation}
with context stream parameters $W_C$.

In addition, an attention inference module is learned to extract attention regions of input, enabling the context encoding stream to focus on the sailent contexts.
Concretely, the attention inference module takes an intermediate feature $X_C$ as input to infer the attention $A \in \mathbb{R}^{H \times W}$, where $H\times W$ is the spatial resolution of the $X_C$.
To make the sum of attention for each pixel to be 1, we spatially normalize the attention $A$ by using the spatial softmax~\cite{sharma2015action} as follows:
\begin{equation}
\hat{A}_{i} = \frac{\mathrm{exp}(A_i)}{\sum_{j} \mathrm{exp}(A_j)},
\end{equation}
where $\hat{A}$ is the attention for context at each pixel $i$ and $j \in \{1, \cdots, H \times W\}$.
Since we temporally aggregate the features using 3D-CNNs, we only normalize the attention weight across spatial axises not temporal axis.
Note that the attention is implicitly learned in an unsupervised manner.
Attention $\hat{A}$ is then applied to the feature $X_C$ to make the attention-boosted feature $\hat{X}_C$ as follows:
\begin{equation}
\bar{X}_C = \hat{A} \odot X_C,
\end{equation}
where $\odot$ is an element-wise multiplication operator.

Specifically, we use five convolution layers to extract intermediate feature volumes $X_C$ followed by BN and ReLU, and 4 max-pooling layers.
All max-pooling layers except for the first layer have $2 \times 2 \times 2$ kernel with stride $2$.
The first pooling layer has kernel size $1 \times 2 \times 2$ similar to facial expression encoding stream.
The number of filters for five convolution layers are 32, 64, 128, and 256, respectively.
In the attention inference module, we use two convolution layers with $3 \times 3 \times 3$ kernels producing 128 and 1 feature channels, followed by BN and ReLU layers.
The final feature $\bar{X}_C$ is spatially averaged in the average-pooling layer.\vspace{-10pt}

\begin{figure}
	\centering
	\renewcommand{\thesubfigure}{}
	\subfigure[(a) input]{\includegraphics[width=0.33\linewidth]{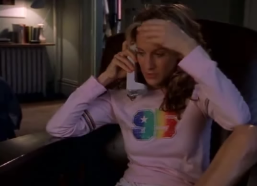}}\hfill
	\subfigure[(b) static model]{\includegraphics[width=0.33\linewidth]{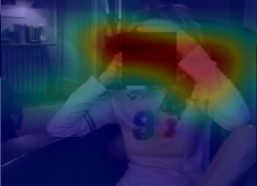}}\hfill
	\subfigure[(c) dynamic model]{\includegraphics[width=0.33\linewidth]{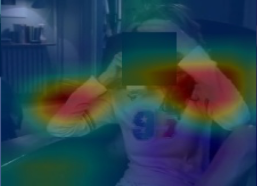}}\hfill
	\caption{Visualization of the attention maps of (b) static and (c) dynamic context encoding models of CAER-Net.}
	\label{fig:3}\vspace{-10pt}
\end{figure}

\paragraph{Static model.}
Dynamic model described above can be simplified for emotion recognition in images. A static model, called CAER-Net-S, takes both a single frame face-cropped image $I_F$ and face-hidden image $I_C$ as input.
In networks, all 3D convolution layers and 3D max-pooling layers are replaced with 2D convolution layers and 2D max-pooling layers, respectively.
Thus, our two types of models can be applied in various environments regardless of the data type.

\figref{fig:3} visualizes the attention maps of static and dynamic models. As expected, our networks both with static and dynamic models localize the context information well, except for the face expression.
By exploiting the temporal connectivity, the dynamic model can localize more sailent regions compared to the static model.

\begin{figure}
	\centering
	\renewcommand{\thesubfigure}{}
	\subfigure{\includegraphics[width=0.33\linewidth]{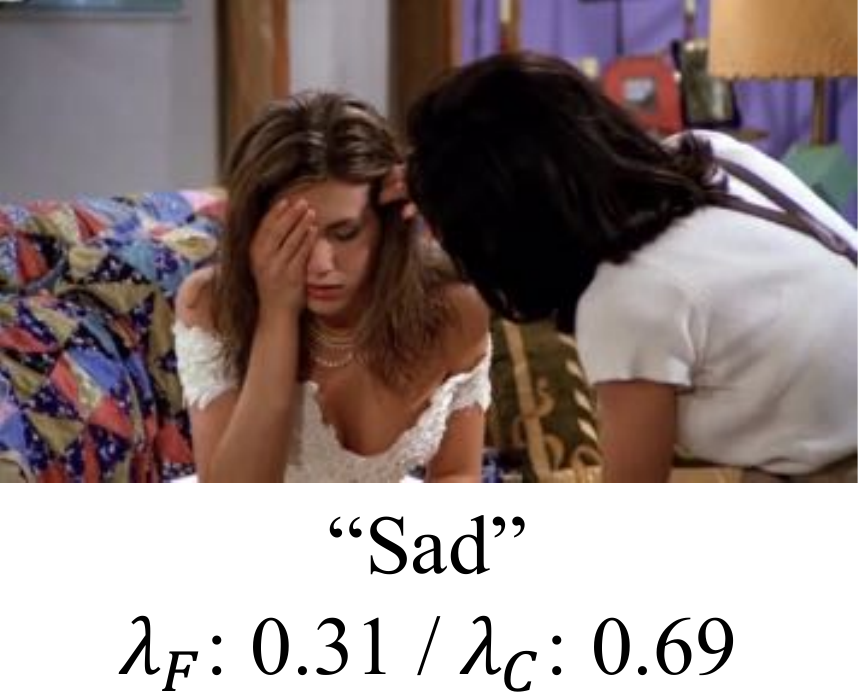}}\hfill
	\subfigure{\includegraphics[width=0.33\linewidth]{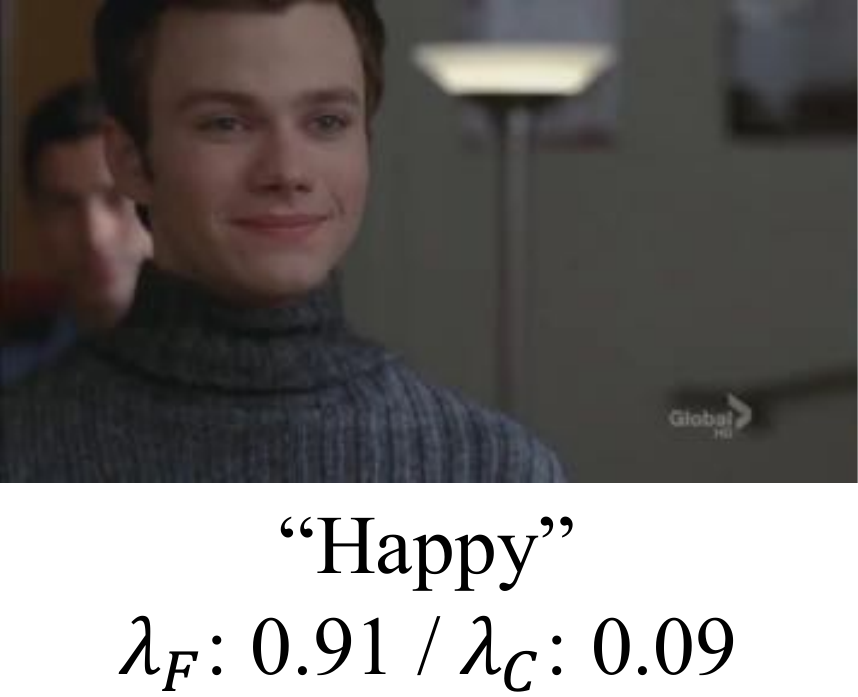}}\hfill
	\subfigure{\includegraphics[width=0.33\linewidth]{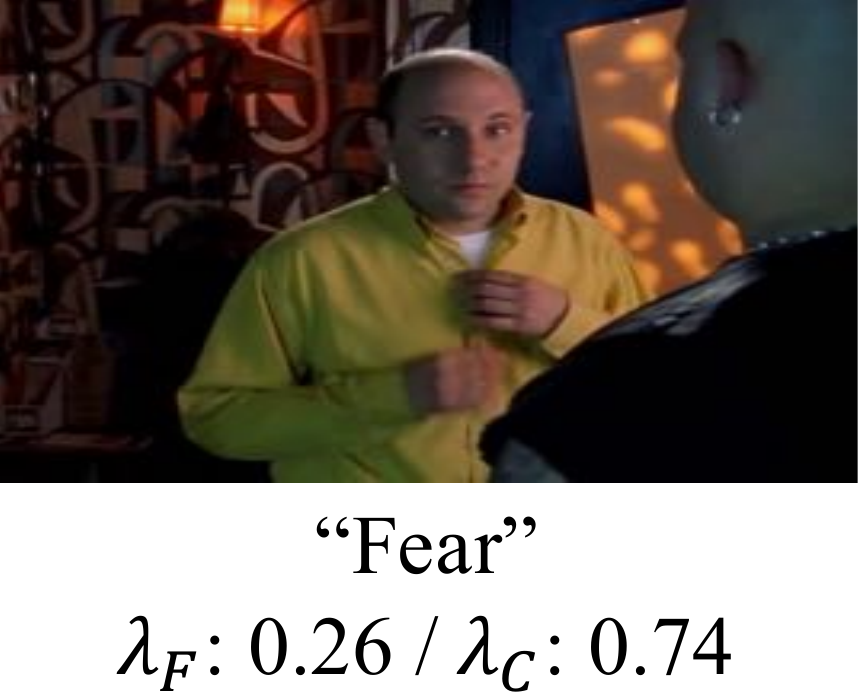}}\hfill \\
	\vspace{-5pt}
	\subfigure{\includegraphics[width=0.33\linewidth]{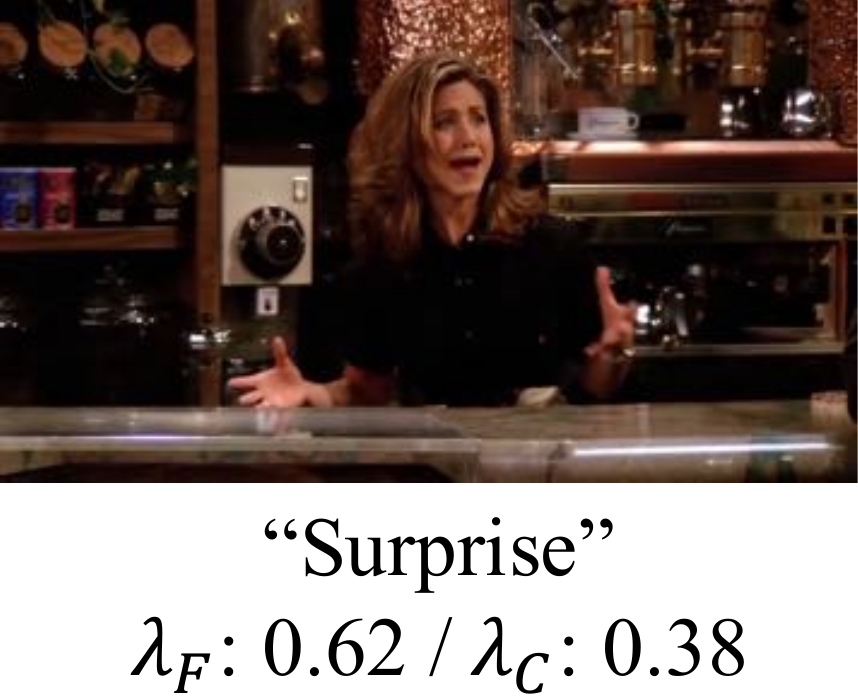}}\hfill
	\subfigure{\includegraphics[width=0.33\linewidth]{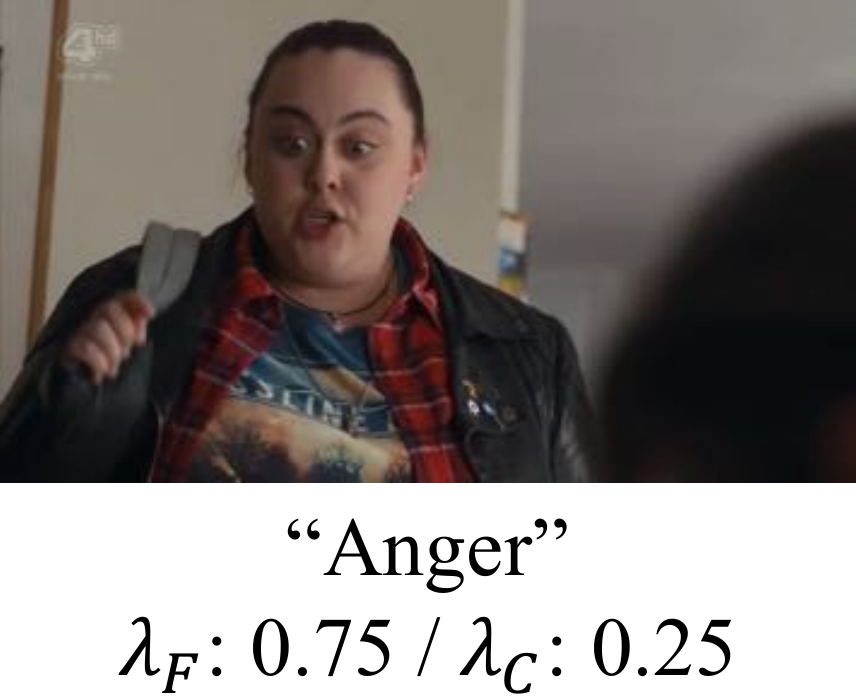}}\hfill
	\subfigure{\includegraphics[width=0.33\linewidth]{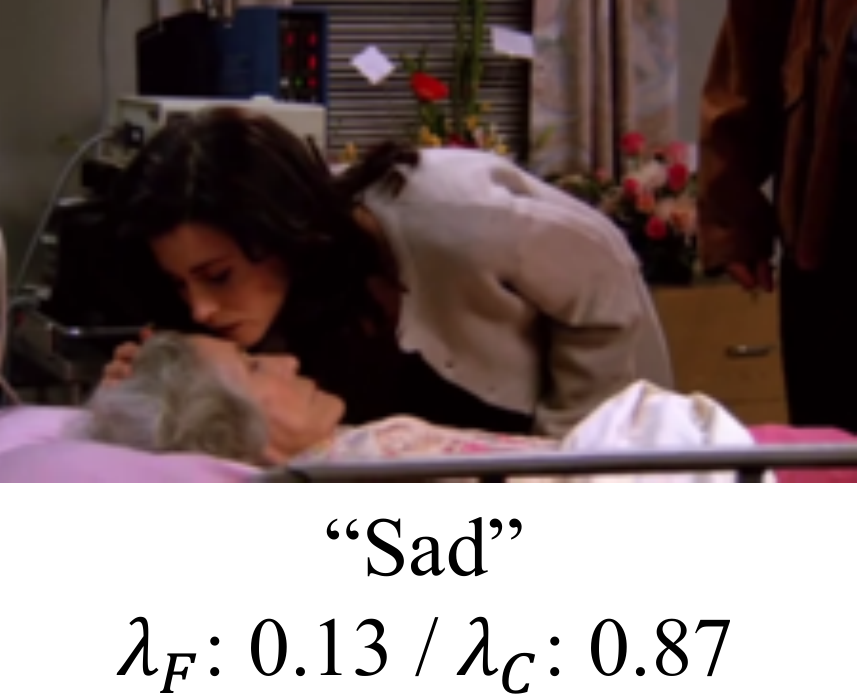}}\hfill \\
	\caption{Some examples of the attention weights, i.e., $\lambda_F$ and $\lambda_C$, in our networks.}
	\label{fig:4}\vspace{-10pt}
\end{figure}

\subsubsection{Adaptive Fusion Networks}\vspace{-5pt}
To recognize the emotion by using the face and context information in a joint manner, the features extracted from two modules should be combined.
However, a direct concatenation of different features~\cite{kosti2017emotion} often fails to provide optimal performance.
To alleviate this limitation, we build the adaptive fusion networks with an attention model for inferring an optimal fusion weight for each feature $X_F$ and $\bar{X}_C$.
The attentions are learned such that $\lambda_F = \mathcal{F}(X_F; W_D)$ and $\lambda_C = \mathcal{F}(\bar{X}_C; W_E)$ with network parameters $W_D$ and $W_E$, respectively. Softmax function make the sum of these attentions to be $1$, \ie, $\lambda_F + \lambda_C = 1$.
\figref{fig:4} shows some examples of the attention weights, i.e., $\lambda_F$ and $\lambda_C$, in CAER-Net. According to contents, the attention weights are adaptively determined to yield an optimal solution.

Unlike methods using the simple concatenation~\cite{kosti2017emotion}, the learned attentions are applied to inputs as
\begin{equation}
X_A = \Pi(X_F \odot \lambda_F, \bar{X}_C \odot \lambda_C),
\end{equation}
where $\Pi$ is a concatenation operator.
We then estimate the final output $y$ for emotion category by classifier:
\begin{equation}
y = \mathcal{F}(X_A; W_G),
\end{equation}
where $W_G$ represents the remainder parameters of the adaptive fusion networks.
\begin{figure}
	\centering
	\renewcommand{\thesubfigure}{}
	\subfigure[]{\includegraphics[width=1.0\linewidth]{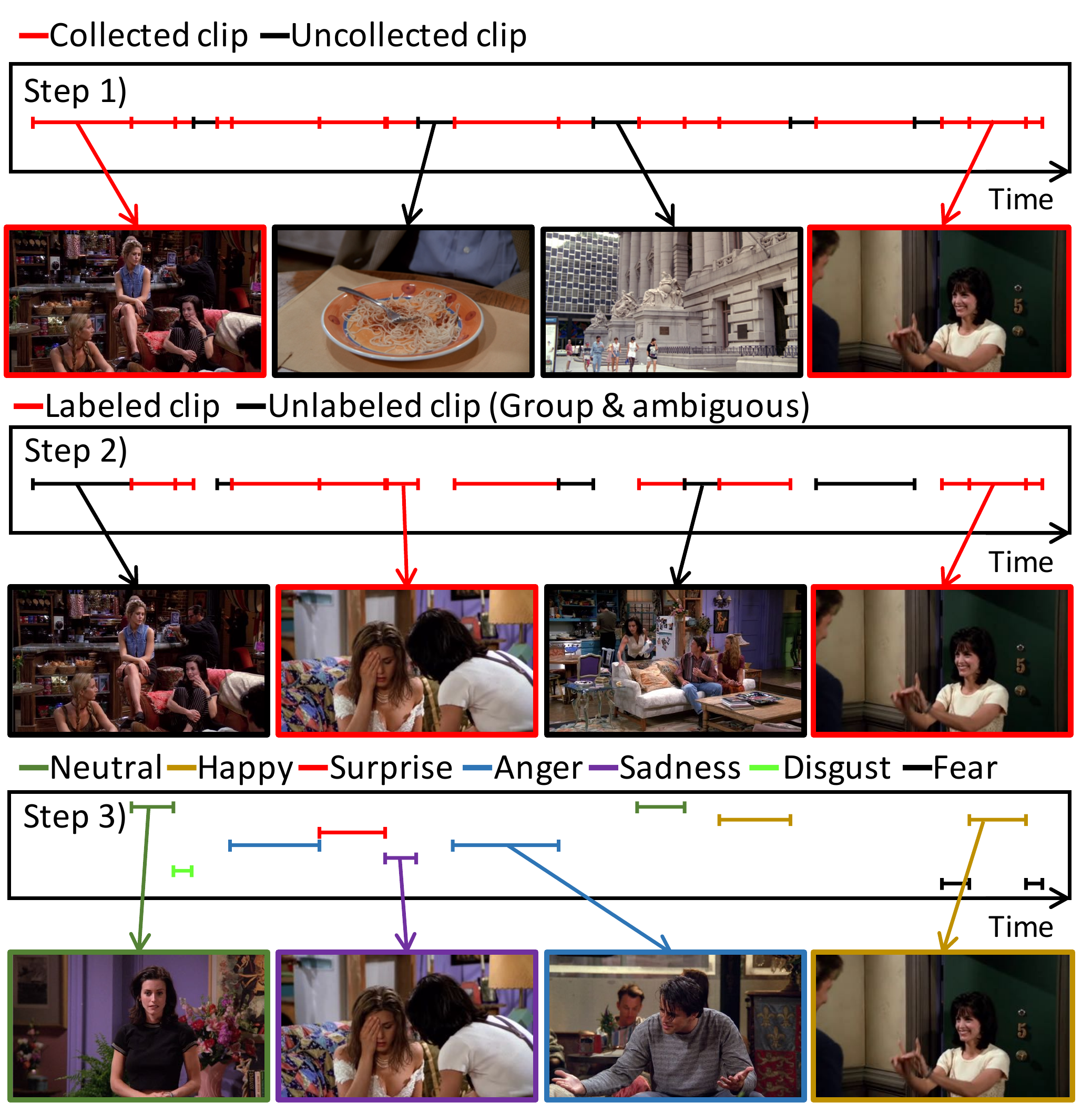}}\\
	\vspace{-10pt}
	\caption{Procedure for building CAER benchmark: we divide the video clips to the shot with shot boundary detection method, and remove face-undetected shots, group-level and ambiguous shots to estimate the emotion. Finally, we annotate the emotion category.}\label{fig:5}\vspace{-10pt}
\end{figure}

\begin{figure*}
	\centering
	\renewcommand{\thesubfigure}{}
	\subfigure[(a) EMOTIC~\cite{kosti2017emotion}]{\includegraphics[height=0.192\textheight]{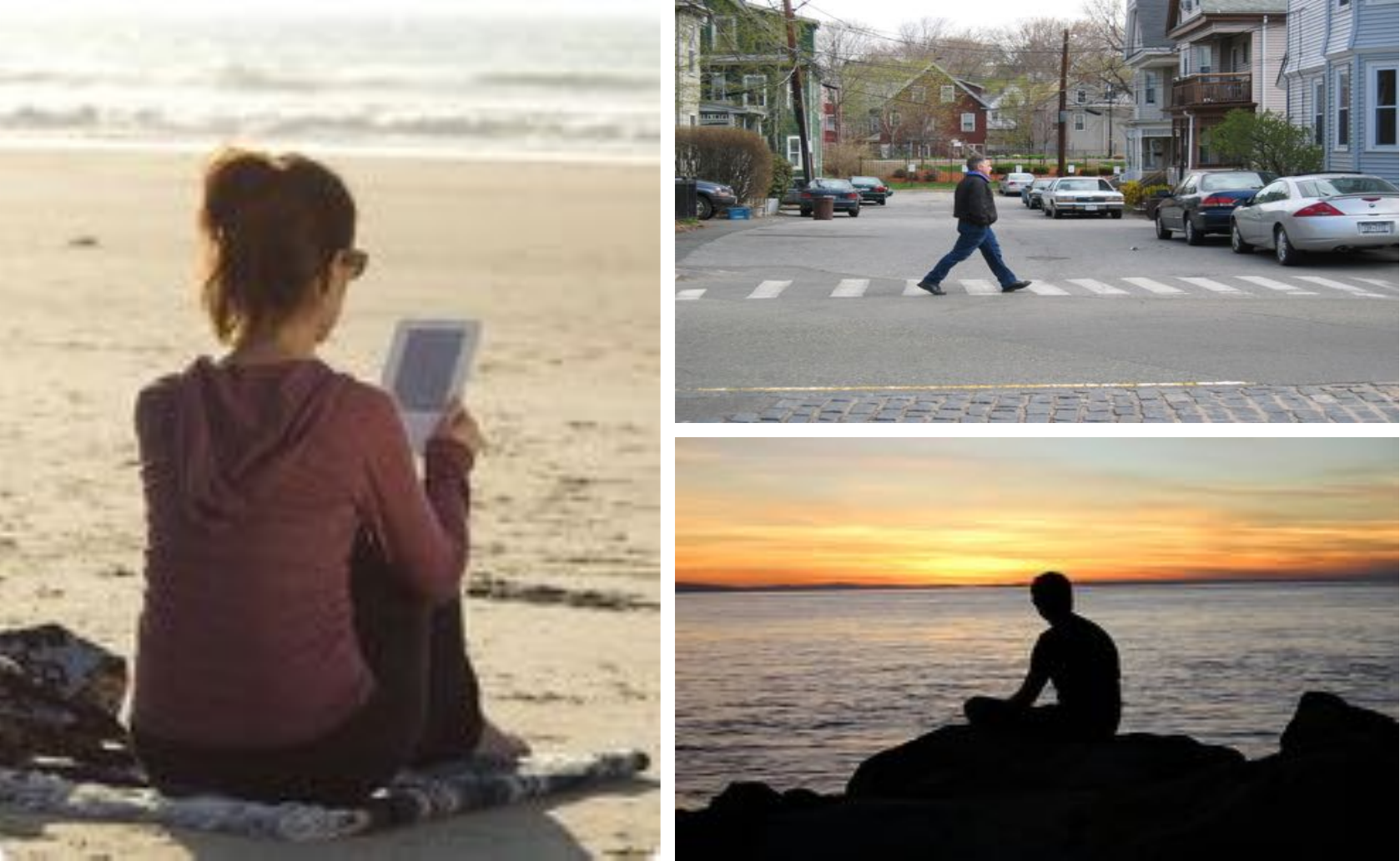}}
	\subfigure[(b) AffectNet~\cite{mollahosseiniaffectnet}]{\includegraphics[height=0.192\textheight]{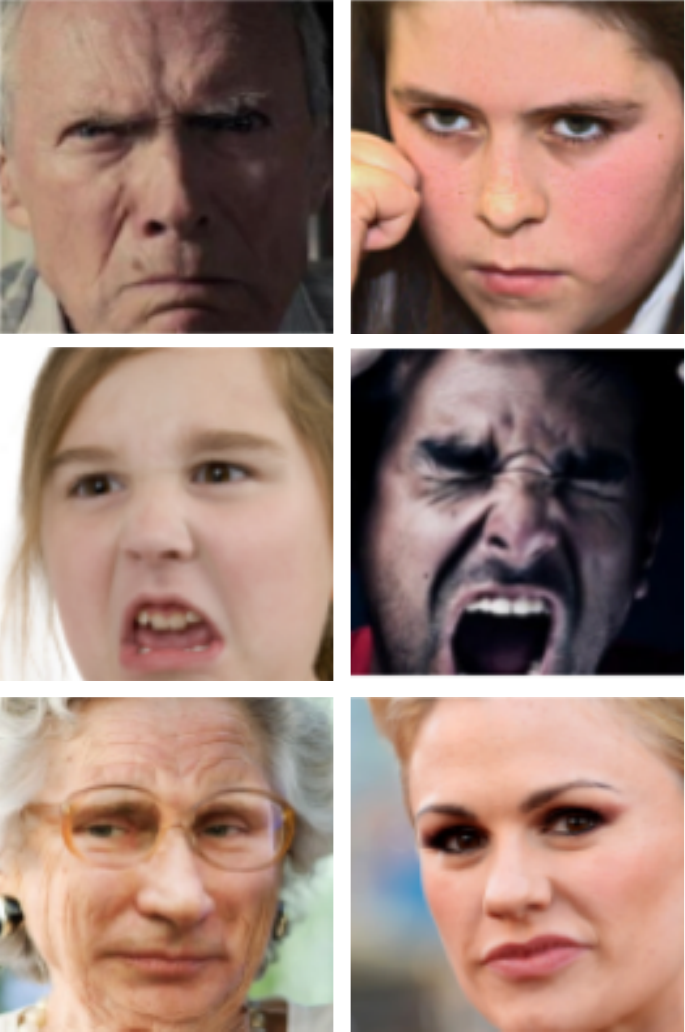}}
	\subfigure[(c) CAER]{\includegraphics[height=0.192\textheight]{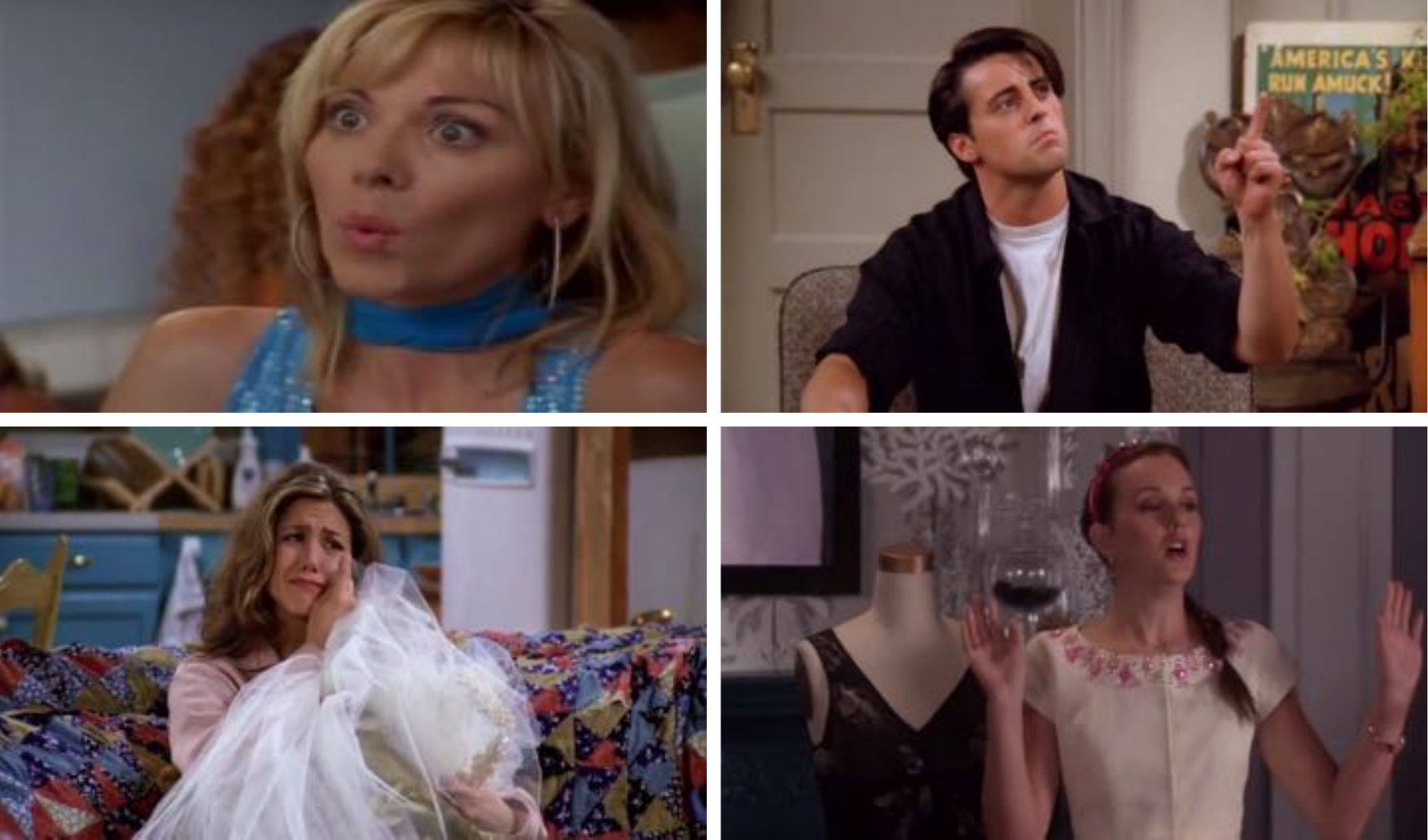}}
	\vspace{-5pt}
	\caption{Examples in the EMOTIC~\cite{kosti2017emotion}, AffectNet~\cite{mollahosseiniaffectnet} and CAER. While EMOTIC includes face-unvisible images to yeild ambiguous emotion recognition, AffectNet includes face-cropped images which have limited to use of context.}
	\label{fig:6}\vspace{-5pt}
\end{figure*}

Specifically, the fusion networks consist of 6 convolution layers with $1 \times 1$ kernels.
The four layers use to produce fusion attention $\lambda_{F}$ and $\lambda_{C}$.
While the intermediate two layers that receive each stream feature as input produce 128 channel feature, the remaining two layers produce 1 channel attention for facial and contextual features.
For the two layers that act as final classifiers, the first convolution layer produces 128 channel feature followed by ReLU and dropout layers to prevent the problem of the network overfitting, and the second convolution layer produces $K$ channel feature to estimated the emotional category.

\section{The CAER Benchmark}\label{sec:4}
Most existing datasets~\cite{goodfellow2013challenges, mollahosseiniaffectnet} have focused on the human facial analysis, and thus they are inappropriate for context-aware emotion recogntion.
In this section, we introduce a benchmark by collecting large-scale video clips from TV shows and annotating them for context-aware emotion recogntion.

\subsection{Annotation}\label{sec:41}
We first collected the video clips from 79 TV shows and then refined them using the shot boundary detector, face detector/tracking and feature clustering~\footnote{https://github.com/pyannote/pyannote-video}. Each video clip was manually annotated with six emotion categories, including ``anger", ``disgust", ``fear", ``happy", ``sad", and ``surprise``, as well as ``neutral".
Six annotators were recruited to assign the emotion category on the 20,484 clips of the initial collection.
Since all the video clips have audio and visual tracks, the annotators labeled them while listening to the audio tracks for more accurate annotations. 
Each clip was evaluated by three different annotators.
The annotation was performed blindly and independently, \ie the annotators were not aware of the other annotator's response.
Importantly, in comparison of existing datasets~\cite{dhall2011acted,kosti2017emotion}, confidence scores were annotated as well as emotion category, which can be thought as the probability of the annotation reliability.
If two more annotators assigned the same emotion categories, the clip was remained in the database.
We also removed the clips which have lower confidence average under the $0.5$.
Finally, 13,201 clips and about 1.1M frames were available. The videos range from short (around 30 frames) to longer clips (more than 120 frames).
The average of sequence length is 90 frames.
In addition, we extracted about 70K static images from CAER to create a static image subset, called CAER-S. The dataset is randomly split into training (70\%), validation (10\%), and testing (20\%) sets.
Overall stage of data acquisition and annotation is illustrated in \figref{fig:5}. \tabref{tab:1} summarizes the number of clips per each cateogry in the CAER benchmark.
\begin{table}
	\begin{center}
		\begin{tabular*}{\linewidth}{l @{\extracolsep{\fill}} ccc}
			\hlinewd{0.8pt}
			Category & \# of clips & \# of frames & \% \tabularnewline
			\hline
			\hline
			Anger		& 1,628	& 139,681	& 12.33	\tabularnewline
			Disgust		& 719	& 59,630	& 5.44\tabularnewline
			Fear		& 514	& 46,441	& 3.89\tabularnewline
			Happy	& 2,726	& 219,377	& 20.64\tabularnewline
			Neutral		& 4,579	& 377,276	& 34.69\tabularnewline
			Sad			& 1,473 & 138,599	& 11.16\tabularnewline
			Surprise	& 1,562 & 126,873	& 11.83\tabularnewline
			\hline
			Total		& 13,201 & 1,107,877 & 100\tabularnewline
			\hlinewd{0.8pt}
		\end{tabular*}
	\end{center}\vspace{-5pt}
	\caption{Amount of video clips in each category on CAER dataset.}
	\label{tab:1}\vspace{-10pt}
\end{table}

\begin{table*}
	\begin{center}
		\begin{tabular}{
				>{\raggedright}m{0.15\linewidth} >{\raggedright}m{0.13\linewidth}
				>{\centering}m{0.13\linewidth} >{\centering}m{0.13\linewidth}
				>{\centering}m{0.13\linewidth} >{\centering}m{0.13\linewidth}}
			\hlinewd{0.8pt}
			{Data type} & {Dataset} & {Amount of data} & {Setting} & {Annotation type} & {Context}\tabularnewline
			\hline
			\hline
			\multirow{3}{*}{{Static (Images)}}
			& EMOTIC~\cite{kosti2017emotion}	& 18,316 images & Web & 26 Categories & \cmark \tabularnewline
			& AffectNet~\cite{mollahosseiniaffectnet}		& 450,000 images & Web & 8 Categories	& \xmark \tabularnewline
			& {CAER-S}		& 70,000 images & TV show & 7 Categories & \cmark \tabularnewline
			\hline
			\multirow{2}{*}{{Dynamic (Videos)}}
			& AFEW~\cite{dhall2012collecting}	& 1,809 clips & Movie & 7 Categories & \xmark \tabularnewline
			& {CAER}		& 13,201 clips & TV show & 7 Categories & \cmark \tabularnewline
			\hlinewd{0.8pt}
		\end{tabular}
	\end{center}
	\vspace{-5pt}
	\caption{Comparison of the CAER with existing emotion recognition datasets such as EMOTIC~\cite{kosti2017emotion}, AffectNet~\cite{mollahosseiniaffectnet}, AFEW~\cite{dhall2012collecting}, and Video Emotion~\cite{jiang2014predicting} datasets. Compared to existing datasets, CAER contains large amount of video clips for context-aware emotion recognition.}
	\label{tab:2}\vspace{-10pt}
\end{table*}

\subsection{Analysis}\label{sec:42}
We compare CAER and CAER-S datasets with other widely used datasets, such as EMOTIC~\cite{kosti2017emotion}, AffectNet~\cite{mollahosseiniaffectnet}, AFEW~\cite{dhall2012collecting}, and Video Emotion datasets~\cite{jiang2014predicting}, as shown in \tabref{tab:2}. According to the data type, the datasets are grouped into the static and dynamic. Even if static databases for facial expression analysis such as AffectNet~\cite{mollahosseiniaffectnet} and FER-Wild~\cite{mollahosseini2016facial} collect a large amount of facial expression images from the web, they have only face-cropped images not including surrounding context.
In addition, EMOTIC~\cite{kosti2017emotion} do not contain human facial images, as exampled in \figref{fig:6}, thus causing subjective and ambiguous labelling from observers. On the other hand, commonly used video emotion recognition datasets had insufficient amount of data than image-based datasets~\cite{jiang2014predicting,kossaifi2017afew}.
Compared to these datasets, the CAER dataset provides the large-scale video clips which are sufficient amount to learn the machine learning algorithms for context-aware emotion recognition.

\section{Experiments}\label{sec:5}
\subsection{Implementation Details}\label{sec:51}
CAER-Net was implemented with PyTorch library~\cite{paszke2017automatic}.
We trained CAER-Net from scratch with learning rate initialized as $5 \times 10^{-3}$ and dropped by a factor of 10 every 4 epochs.
CAER-Net was learned with the cross-entropy loss function~\cite{kim2019unified} with ground-truth emotion labels with batch size to 32.
As CAER dataset has various length of videos, we randomly extracted single non-overlapped consecutive 16 frame clips from every training video which sampled at 10 frames per second.
While the clips of facial $V_F$ are resized to have the frame size of $96 \times 96$, the clips of contextual parts $V_C$ are resized to have the frame size of $128 \times 171$ and randomly cropped into $112 \times 112$ at training stage.
We also trained static model of CAER-Net-S with CAER-S dataset with the input size of $224 \times 224$.
To reduce the effects of overfitting, we employed the dropout scheme with the ratio of 0.5 between $1 \times 1$ convolution layers, and data augmentation schemes such as flips, contrast, and color changes.
At testing phase, we used a single center crop per contextual parts clips.
For video predictions, we split a video into 16 frame clips with a 8 frame overlap between two consecutive clips then average clip predictions of all clips.

\subsection{Experimental Settings}\label{sec:51}
We evaluated CAER-Net on the CAER and AFEW dataset~\cite{dhall2011acted}, respectively.
For evaluation of the proposed networks quantitatively, we measured the emotion recognition performance by classification accuracy as used in~\cite{dhall2016emotiw}.
We reproduced four classical deep network architectures before the fully-connected layers, including AlexNet~\cite{krizhevsky2012imagenet}, VGGNet~\cite{simonyan2014very}, ResNet~\cite{he2016deep}, and C3D~\cite{tran2015learning}, as the baseline methods.
We adopt two fully-connected layers as classifiers for the baseline methods.
We initialized the feature extraction modules of all the baselines using pretrained models from two large-scale classification datasets such as ImageNet~\cite{deng2009imagenet} and Sports-1M~\cite{karpathy2014large}, and fine-tuned whole networks on CAER benchmark.
We trained all parameters of learning rate $10^{-4}$ for fine-tuned models.

\begin{table}
	\begin{center}
		\begin{tabular}{
				>{\raggedright}m{0.22\linewidth} >{\centering}m{0.07\linewidth}
				>{\centering}m{0.07\linewidth} >{\centering}m{0.08\linewidth}
				>{\centering}m{0.08\linewidth} >{\centering}m{0.16\linewidth}}
			\hlinewd{0.8pt}
			Methods & w/F  & w/C  & w/cA & w/fA & Acc. (\%) \tabularnewline
			\hline
			\hline
			\multirow{3}{*}{CAER-Net-S} & \cmark	&  & & & 70.09\tabularnewline
			& 			&	\cmark	& \cmark& & 65.65 \tabularnewline
			& \cmark	& 	\cmark	& \cmark& \cmark& 73.51 \tabularnewline 
			\hline
			\multirow{6}{*}{CAER-Net} & \cmark	&				&&& 74.13\tabularnewline
			& 			&	\cmark	& \cmark && 71.94 \tabularnewline
			& \cmark	&	\cmark	&  && 74.36 \tabularnewline
			& \cmark	&	\cmark	& \cmark&& 74.94 \tabularnewline
			& \cmark	&	\cmark	& & \cmark& 75.57 \tabularnewline
			& \cmark	& 	\cmark	& \cmark & \cmark& {77.04} \tabularnewline 
			\hlinewd{0.8pt}
		\end{tabular}
	\end{center}
	\vspace{-5pt}
	\caption{Ablation study of CAER-Net-S and CAER-Net on the CAER-S and CAER datasets, respectively. `F', `C', `cA', and `fA' denote face encoding stream, context encoding stream, context attention module and fusion attention module, respectively.}\label{tab:3}\vspace{-10pt}
\end{table}

\subsection{Results on the CAER dataset}\label{sec:53}
\paragraph{Ablation study.}
We analyzed CAER-Net-S and CAER-Net with ablation studies as varying the combination of different inputs such as cropped face and context, and attention modules such as context and fusion attention modules.
For all those experiments, CAER-Net-S and CAER-Net were trained and tested on the CAER-S and CAER datasets, respectively.
For quantitative analysis of ablation study, we examined the classification accuracy on the CAER benchmark as shown in \tabref{tab:3}. The results show that the best result can be obtained when both the face and context are used as inputs.
As our baseline, CAER-Net w/F that considers facial expression only for emotion recognition provides the accuracy 74.13 $\%$. Compared to this, our CAER-Net that fully makes use of both face and context shows the best performance.
When we compared the static and dynamic models, CAER-Net shows 3.53 $\%$ improvement than CAER-Net-S, which shows the importance to consider the temporal dynamic inputs for context-aware emotion recognition.

\begin{figure}
	\centering
	\renewcommand{\thesubfigure}{}
	\subfigure[(a) CAER-Net w/F]{\includegraphics[width=0.5\linewidth]{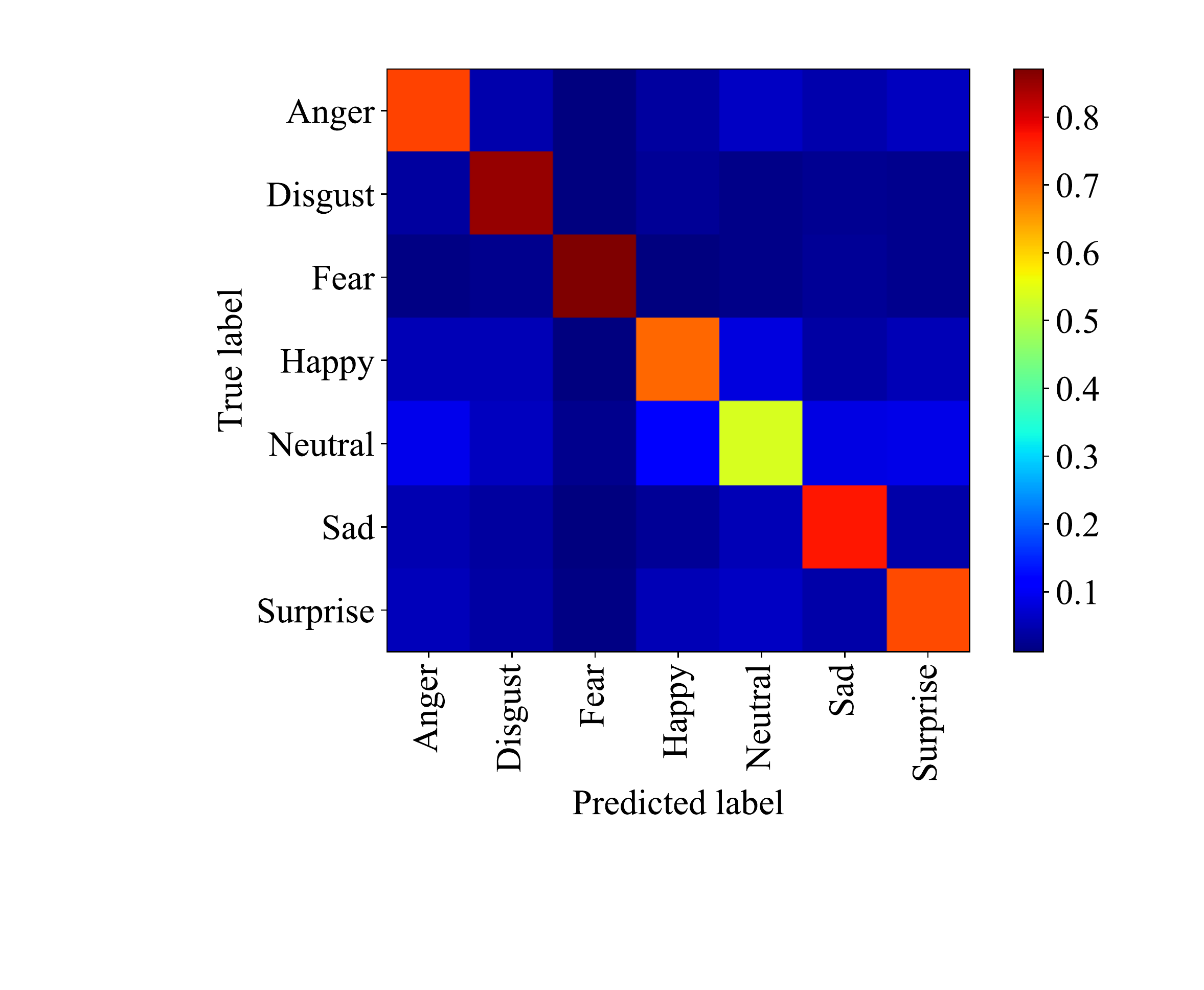}}\hfill
	\subfigure[(b) CAER-Net]{\includegraphics[width=0.5\linewidth]{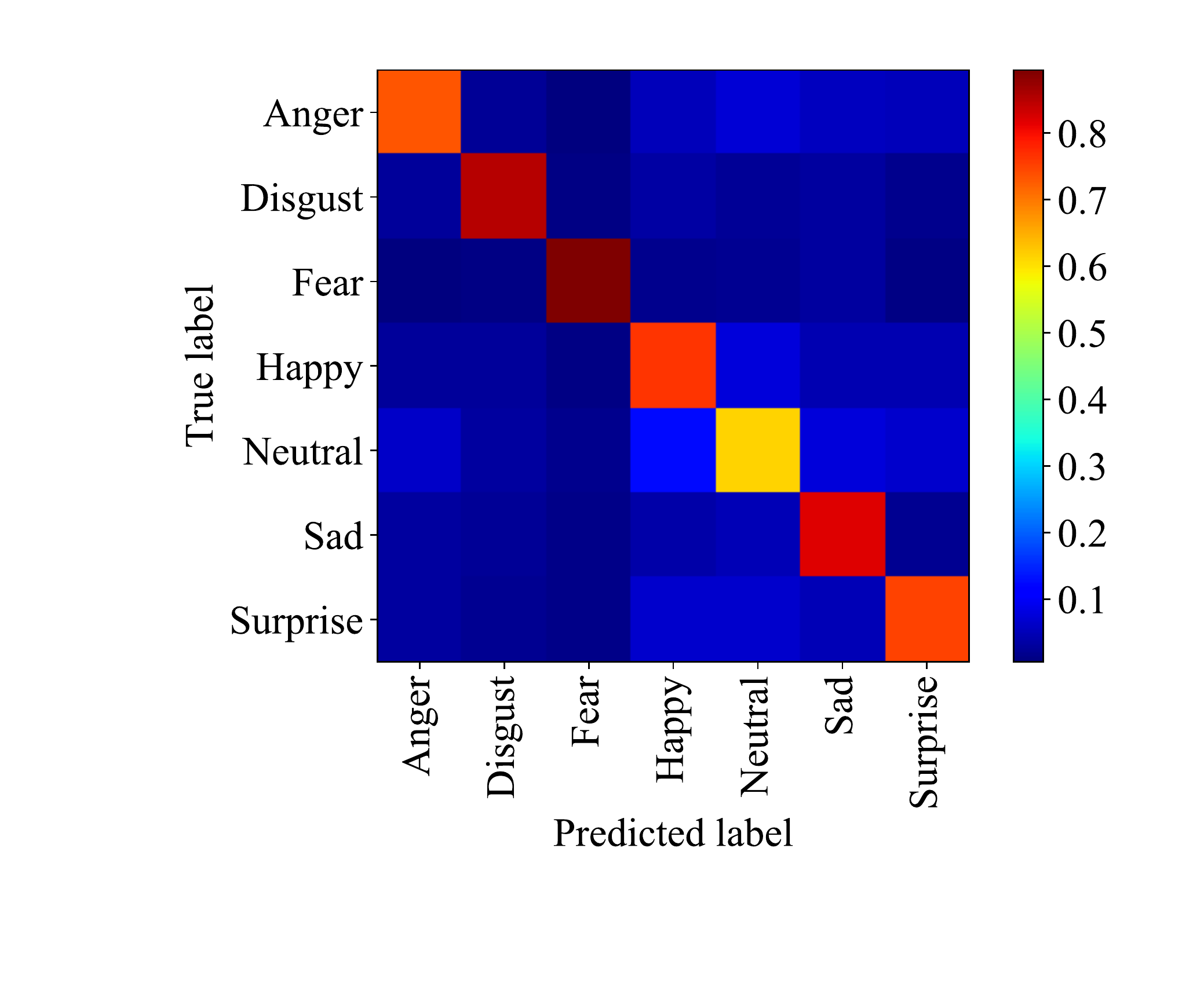}}\hfill\\
	\caption{Confusion matrix of CAER-Net with face stream only and with face and context streams on the CAER benchmark.}
	\label{fig:7}\vspace{-8pt}
\end{figure}

\begin{figure}
	\centering
	\renewcommand{\thesubfigure}{}
	\subfigure{\includegraphics[width=0.248\linewidth]{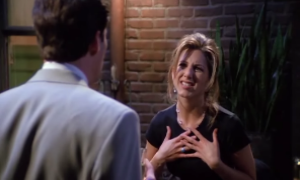}}\hfill
	\subfigure{\includegraphics[width=0.248\linewidth]{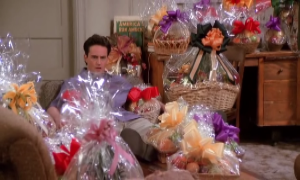}}\hfill
	\subfigure{\includegraphics[width=0.248\linewidth]{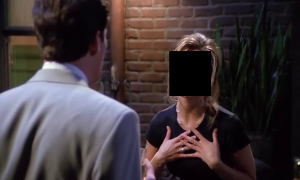}}\hfill
	\subfigure{\includegraphics[width=0.248\linewidth]{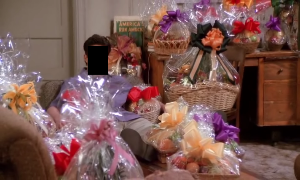}}\hfill\\
    \vspace{-10pt}
	\subfigure{\includegraphics[width=0.248\linewidth]{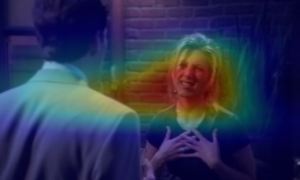}}\hfill
	\subfigure{\includegraphics[width=0.248\linewidth]{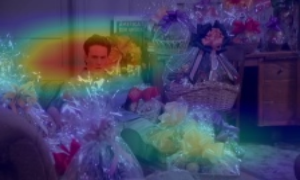}}\hfill
	\subfigure{\includegraphics[width=0.248\linewidth]{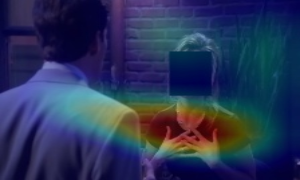}}\hfill
	\subfigure{\includegraphics[width=0.248\linewidth]{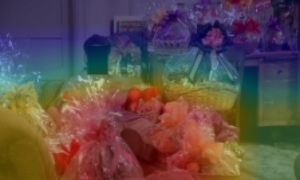}}\hfill\\
    \vspace{-10pt}
	\subfigure[(a)]{\includegraphics[width=0.248\linewidth]{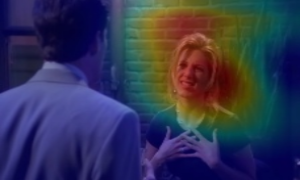}}\hfill
	\subfigure[(b)]{\includegraphics[width=0.248\linewidth]{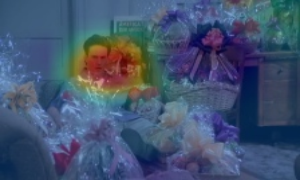}}\hfill
	\subfigure[(c)]{\includegraphics[width=0.248\linewidth]{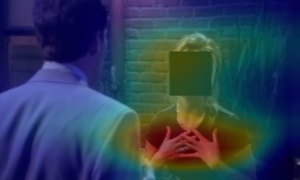}}\hfill
	\subfigure[(d)]{\includegraphics[width=0.248\linewidth]{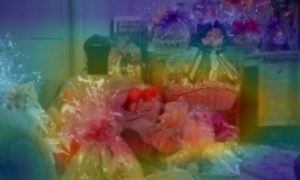}}\hfill\\
	\vspace{-3pt}
	\caption{Visualization of the attention: (from top to bottom) inputs, attention maps of CAER-Net-S and CAER-Net. (a) and (b) are results of ablation study without hiding the face during training, (c) and (d) with hiding the face.}
	\label{fig:10}\vspace{-10pt}
\end{figure}

\figref{fig:7} demonstrates the confusion matrix of CAER-Net w/F and CAER-Net, which also verify that compared to the model that only focuses on facial stream only, a joint model that considers facial stream and context stream simultaneously can highly boost the emotion recognition performance. Happy and neutral accuracies were increased by 7.48\% and 5.65\%, respectively, which clearly shows that context information helps distinguishing these two categories rather than only using facial expression.
Finally, we conducted an ablation study for the context attention module. First of all, when we trained CAER-Net-S and CAER-Net without hiding the face, they tended to focus on the most discriminative parts only (\ie, faces) as depicted in the preceding two columns \figref{fig:10}. Secondly, we conducted another experiment on \emph{actionless} frames as depicted in the second and last columns. As shown in the last two columns \figref{fig:10}, both CAER-Net-S and CAER-Net attend to not only ``things that move" but also the salient scene that can be an emotion signals.
To summarize, our context encoding stream enables the networks to attend salient context that boost performance for both images and videos.

\begin{figure}
	\centering
	\renewcommand{\thesubfigure}{}
	\subfigure[]{\includegraphics[width=1\linewidth]{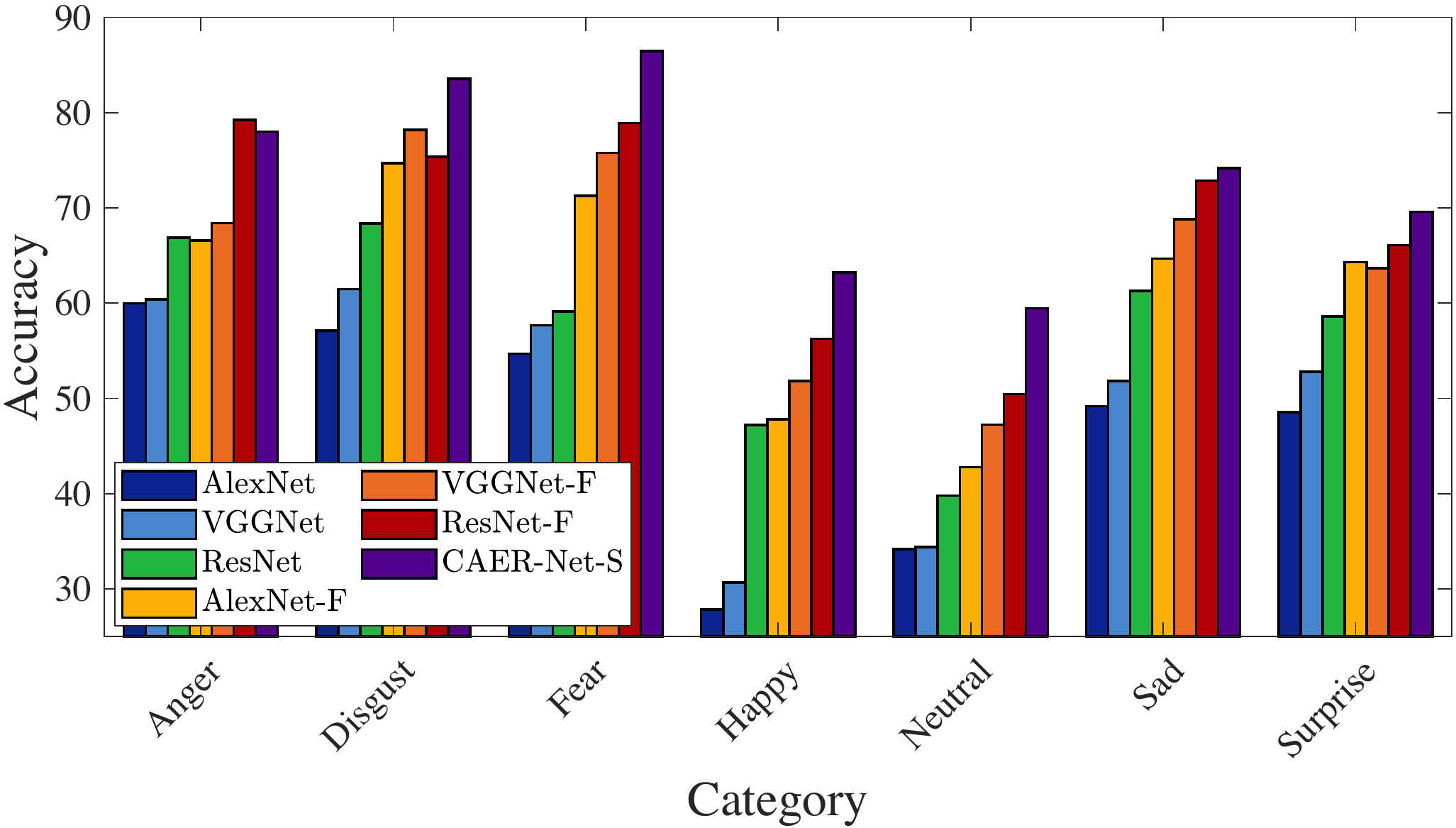}}\\
	\vspace{-13pt}
	\caption{Quantitative evaluation of CAER-Net-S in comparison to baseline methods on each category in the CAER-S benchmark.}\vspace{-5pt}\label{fig:8}
\end{figure}

\begin{figure*}
	\centering
	\renewcommand{\thesubfigure}{}
	\subfigure{\includegraphics[width=0.164\linewidth]{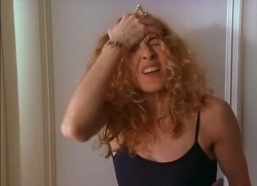}}\hfill
	\subfigure{\includegraphics[width=0.164\linewidth]{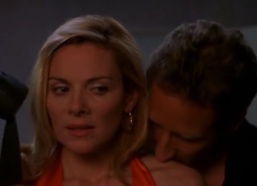}}\hfill
	\subfigure{\includegraphics[width=0.164\linewidth]{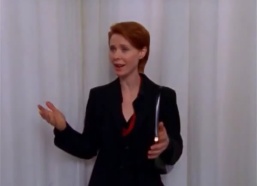}}\hfill
	\subfigure{\includegraphics[width=0.164\linewidth]{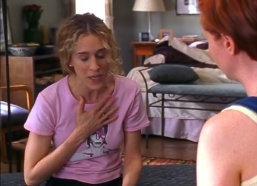}}\hfill
	\subfigure{\includegraphics[width=0.164\linewidth]{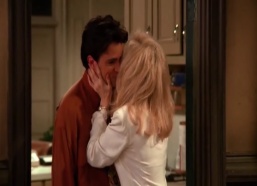}}\hfill
	\subfigure{\includegraphics[width=0.164\linewidth]{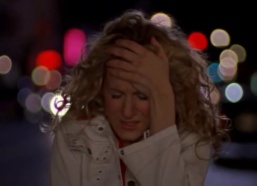}}\hfill\\
	\vspace{-10pt}
	\subfigure{\includegraphics[width=0.164\linewidth]{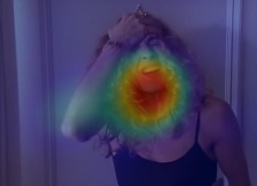}}\hfill
	\subfigure{\includegraphics[width=0.164\linewidth]{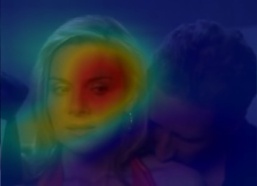}}\hfill
	\subfigure{\includegraphics[width=0.164\linewidth]{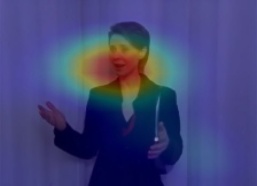}}\hfill
	\subfigure{\includegraphics[width=0.164\linewidth]{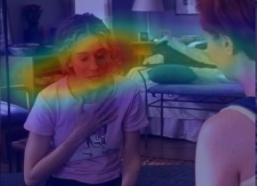}}\hfill
	\subfigure{\includegraphics[width=0.164\linewidth]{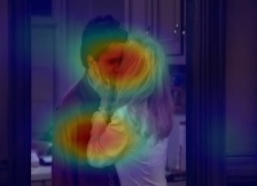}}\hfill
	\subfigure{\includegraphics[width=0.164\linewidth]{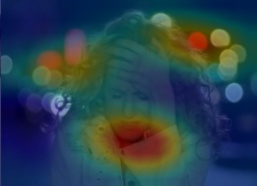}}\hfill\\
	\vspace{-10pt}
	\subfigure{\includegraphics[width=0.164\linewidth]{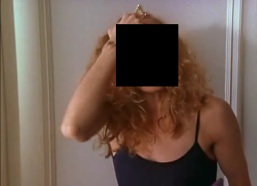}}\hfill
	\subfigure{\includegraphics[width=0.164\linewidth]{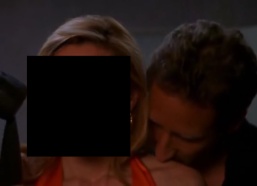}}\hfill
	\subfigure{\includegraphics[width=0.164\linewidth]{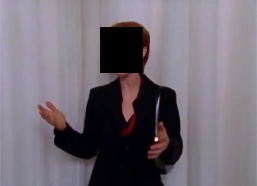}}\hfill
	\subfigure{\includegraphics[width=0.164\linewidth]{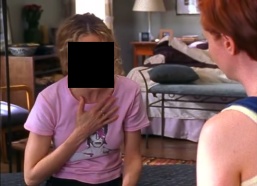}}\hfill
	\subfigure{\includegraphics[width=0.164\linewidth]{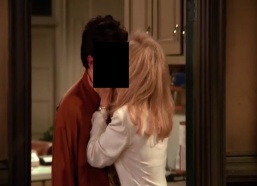}}\hfill
	\subfigure{\includegraphics[width=0.164\linewidth]{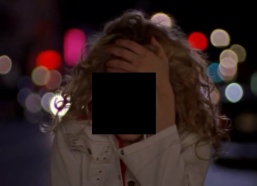}}\hfill \\
	\vspace{-10pt}
	\subfigure[(a) ``Disgust"]{\includegraphics[width=0.164\linewidth]{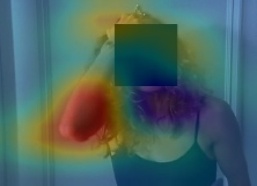}}\hfill
	\subfigure[(b) ``Fear"]{\includegraphics[width=0.164\linewidth]{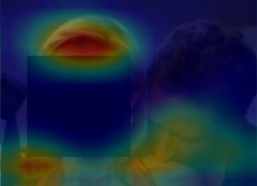}}\hfill
	\subfigure[(c) ``Surprise"]{\includegraphics[width=0.164\linewidth]{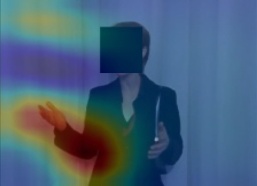}}\hfill
	\subfigure[(d) ``Sad"]{\includegraphics[width=0.164\linewidth]{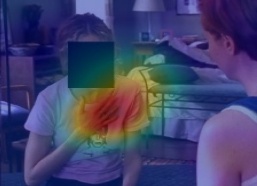}}\hfill
	\subfigure[(e) ``Happy"]{\includegraphics[width=0.164\linewidth]{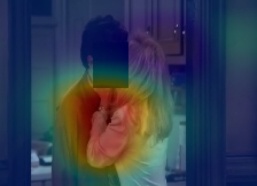}}\hfill
	\subfigure[(f) ``Fear"]{\includegraphics[width=0.164\linewidth]{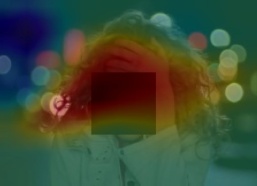}}\hfill\\
	\caption{Visualization of learned attention maps in CAER-Net-S: (from top to bottom) inputs, attention maps of CAM~\cite{zhou2016learning}, inputs of context encoding stream, attention maps in context encoding stream. Note that red color indicates attentive regions and blue color indicates suppressed regions. Best viewed in color.}\vspace{-10pt}
	\label{fig:9}
\end{figure*}

\vspace{-10pt}
\paragraph{Comparison to baseline methods.}
In \figref{fig:8} and \tabref{tab:4}, we evaluated CAER-Net-S with baseline 2D CNNs based approaches.
The standard networks including AlexNet~\cite{krizhevsky2012imagenet}, VGGNet~\cite{simonyan2014very}, and ResNet~\cite{he2016deep} pretrained with ImageNet were reproduced for comparison with CAER-Net-S. In addition, we also fine-tuned these networks on the CAER-S dataset.
Compared to these baseline methods, our CAER-Net-S improves the classification performance than fine-tuned ResNet by 5.05$\%$. Moreover, CAER-Net-S consistently performs favorably against baseline deep networks on each category in the CAER-S benchmark, which illustrates that CAER-Net can learn more discriminative representation for this task.
In addition, we evaluated CAER-Net with a baseline 3D CNNs based approach in \tabref{tab:5}. Compared to C3D~\cite{tran2015learning}, our CAER-Net has shown the state-of-the-art performance on the CAER benchmark.

\begin{table}[!t]
	\begin{center}
		\begin{tabular}{
				>{\raggedright}m{0.5\linewidth} >{\centering}m{0.16\linewidth}}
			\hlinewd{0.8pt}
			Methods & Acc. (\%)\tabularnewline
			\hline
			\hline
			ImageNet-AlexNet~\cite{krizhevsky2012imagenet}  & 47.36\tabularnewline
			ImageNet-VGGNet~\cite{simonyan2014very} & 49.89\tabularnewline
			ImageNet-ResNet~\cite{he2016deep} & 57.33\tabularnewline
			\hline
			Fine-tuned AlexNet~\cite{krizhevsky2012imagenet} & 61.73\tabularnewline
			Fine-tuned VGGNet~\cite{simonyan2014very} & 64.85\tabularnewline
			Fine-tuned ResNet~\cite{he2016deep} & 68.46\tabularnewline
			\hline
			CAER-Net-S & 73.51\tabularnewline
			\hlinewd{0.8pt}
		\end{tabular}
	\end{center}
	\vspace{-5pt}
	\caption{Quantitative evaluation of CAER-Net-S in comparison to baseline methods on the CAER-S benchmark .}\vspace{-10pt}\label{tab:4}
\end{table}

\begin{table}[!t]
	\begin{center}
		\begin{tabular}{
				>{\raggedright}m{0.5\linewidth} >{\centering}m{0.16\linewidth}}
			\hlinewd{0.8pt}
			Methods & Acc. (\%)\tabularnewline
			\hline
			\hline
			Sports-1M-C3D~\cite{tran2015learning} & 66.38 \tabularnewline
			Fine-tuned C3D~\cite{tran2015learning} & 71.02 \tabularnewline
			\hline
			CAER-Net &77.04 \tabularnewline
			\hlinewd{0.8pt}
		\end{tabular}
	\end{center}
	\vspace{-5pt}
	\caption{Quantitative evaluation of CAER-Net in comparison to C3D~\cite{tran2015learning} on the CAER benchmark .}\vspace{-5pt}\label{tab:5}
\end{table}

Finally, \figref{fig:9} shows the qualitative results with learned attention maps obtained by CAM~\cite{zhou2016learning} with fine-tuned VGGNet and in context encoding stream of CAER-Net-S. Note that images in \figref{fig:9} were correctly classified to ground-truth emotion categories both with fine-tuned VGGNet and CAER-Net-S. Unlike CAM~\cite{zhou2016learning} that only considers facial expressions, the attention mechanism in CAER-Net-S localizes context information well that can boost the emotion recognition performance in a context-aware manner.

\subsection{Results on the AFEW dataset}\label{sec:52}\vspace{-3pt}
We conducted an additional experiment to verify the effectiveness of the CAER dataset compared to the AFEW dataset~\cite{dhall2011acted}.
When we trained CAER-Net on the combination of CAER and AFEW datasets, the highly improvement was attained. It demonstrates that CAER dataset could be complement data distribution of the AFEW dataset. It should be noted that Fan \etal~\cite{fan2018video} has shown the better performance, they are formulated the networks with the ensemble of various networks to maximize the performance in EmotiW challenge. Unlike this, we focused on investigating how context information helps to improve the emotion recognition performance. For this purpose, we choice shallow architecture rather than Fan~\etal~\cite{fan2018video}.
If the face encoding stream adopt more complicated networks such Fan \etal~\cite{fan2018video}, the performance of CAER-Net also will be highly boosted. We reserve this as further works.\vspace{-5pt}

\begin{table}[!t]
	\begin{center}
		\begin{tabular}{
				>{\raggedright}m{0.398\linewidth}
				>{\centering}m{0.29\linewidth} >{\centering}m{0.16\linewidth}}
			\hlinewd{0.8pt}
			Methods & Training data& Acc. (\%)\tabularnewline
			\hline
			\hline
			VielZeuf \etal~\cite{vielzeuf2017temporal} w/F & FER+AFEW & 48.60 \tabularnewline
			Fan \etal~\cite{fan2016video} w/F & FER+AFEW & 48.30	\tabularnewline
			Hu \etal~\cite{hu2017learning} w/F & AFEW & 42.55	\tabularnewline
			Fan \etal~\cite{fan2018video} w/F & FER+AFEW & 57.43	\tabularnewline
			\hline
			CAER-Net w/F & AFEW  & 41.86 \tabularnewline
			CAER-Net & CAER  & 38.65 \tabularnewline
			CAER-Net & AFEW  & 43.12 \tabularnewline
			CAER-Net & CAER+AFEW  & 51.68 \tabularnewline
			\hlinewd{0.8pt}
		\end{tabular}
	\end{center}
	\vspace{-5pt}
	\caption{Quantitative evaluation of CAER-Net on the AFEW~\cite{dhall2011acted} benchmark, as varying training datasets.}\vspace{-10pt}\label{tab:6}
\end{table}

\section{Conclusion}\label{sec:6}
We presented CAER-Net that jointly exploits human facial expression and context for context-aware emotion recognition.
The key idea of this approach is to seek sailent context information by hiding the facial regions with an attention mechanism, and utilize this to estimate the emotion from contexts, as well as the facial information together.
We also introduced the CAER benchmark that is more appropriate for context-aware emotion recognition than existing benchmarks both qualitatively and quantitatively.
We hope that the results of this study will facilitate further advances in context-aware emotion recognition and its related tasks.

{\small
	\bibliographystyle{ieee_fullname}
	\bibliography{egbib}
}

\end{document}